\documentclass{bmvc2k}
\usepackage{xcolor}

\usepackage{mathtools}
\usepackage{hyperref}
\usepackage[ruled,vlined,linesnumbered]{algorithm2e}
\usepackage{wrapfig}
\usepackage{amsmath}
\usepackage{amssymb}
\newcommand\blfootnote[1]{%
  \begingroup
  \renewcommand\thefootnote{}\footnote{#1}%
  \addtocounter{footnote}{-1}%
  \endgroup
}

\renewcommand{\thefootnote}{\fnsymbol{footnote}}

\title{An Adaptive Supervision Framework for \\
Active Learning in Object Detection}

\addauthor{Sai Vikas Desai\textsuperscript{$\dagger$}}{cs17mtech11011@iith.ac.in}{1}
\addauthor{Akshay L Chandra\textsuperscript{$\dagger$}}{akshaychandra@iith.ac.in}{1}
\addauthor{Wei Guo}{guowei@isas.a.u-tokyo.ac.jp}{2}
\addauthor{Seishi Ninomiya}{snino@isas.a.u-tokyo.ac.jp}{2}
\addauthor{Vineeth N Balasubramanian}{vineethnb@iith.ac.in}{1}
\addinstitution{
  Indian Institute of Technology Hyderabad, \\ Kandi, 502285, India
}
\addinstitution{
  Graduate School of Agricultural \\and Life Sciences, \\The University of Tokyo, \\Tokyo 1880002, Japan
}

\runninghead{Desai \etal}{Adaptive Supervision for Active Learning}

\def\etal{\emph{et al}\bmvaOneDot}

\begin{document}
\maketitle
\blfootnote{
\hspace{-19pt} $\dagger$Both authors have equally contributed.}
\vspace{-15pt}
\begin{abstract}
Active learning approaches in computer vision generally involve querying strong labels for data. However, previous works have shown that weak supervision can be effective in training models for vision tasks while greatly reducing annotation costs. Using this knowledge, we propose an adaptive supervision framework for active learning and demonstrate its effectiveness on the task of object detection. Instead of directly querying bounding box annotations (strong labels) for the most informative samples, we first query weak labels and optimize the model. Using a switching condition, the required supervision level can be increased. Our framework requires little to no change in model architecture. Our extensive experiments show that the proposed framework can be used to train good generalizable models with much lesser annotation costs than the state of the art active learning approaches for object detection. 
\end{abstract}

\vspace{-12pt}
\section{Introduction}
\label{sec:intro}
\vspace{-5pt}
State-of-the-art performance of deep neural networks in computer vision tasks such as object detection and semantic segmentation has been largely achieved using fully supervised learning methods \cite{maskrcnn,deeplab}, which demand large amounts of strongly annotated data. However, it is known that obtaining labels for vast amounts of data is expensive and time-consuming. In this work, we focus on the problem of training efficient object detectors while minimizing the required annotation effort.

Active learning has been shown to be efficient in reducing labeled data requirement for image classification \cite{Sener2018ActiveLF,Gal2017DeepBA,Wang_gupta,al_gaussian,adaptive_al}. However, fewer efforts have been proposed to attempt active learning for object detection using deep neural networks \cite{Kao2018LocalizationAwareAL,brust,VinayNamboodiri2018}. In these approaches, an oracle is asked to provide accurate bounding box labels for the most informative set of images, which are selected by the corresponding methodology. These methods mostly vary by the nature of the methodology used to choose the query images, or in case of object detection, by the nature of the underlying object detection framework. In this work, we propose a highly effective approach to leverage weak supervision for active learning in object detection. 

\begin{figure}
\begin{tabular}{cc}
\hspace{+20pt}
\bmvaHangBox{\includegraphics[width=5cm, height=4cm]{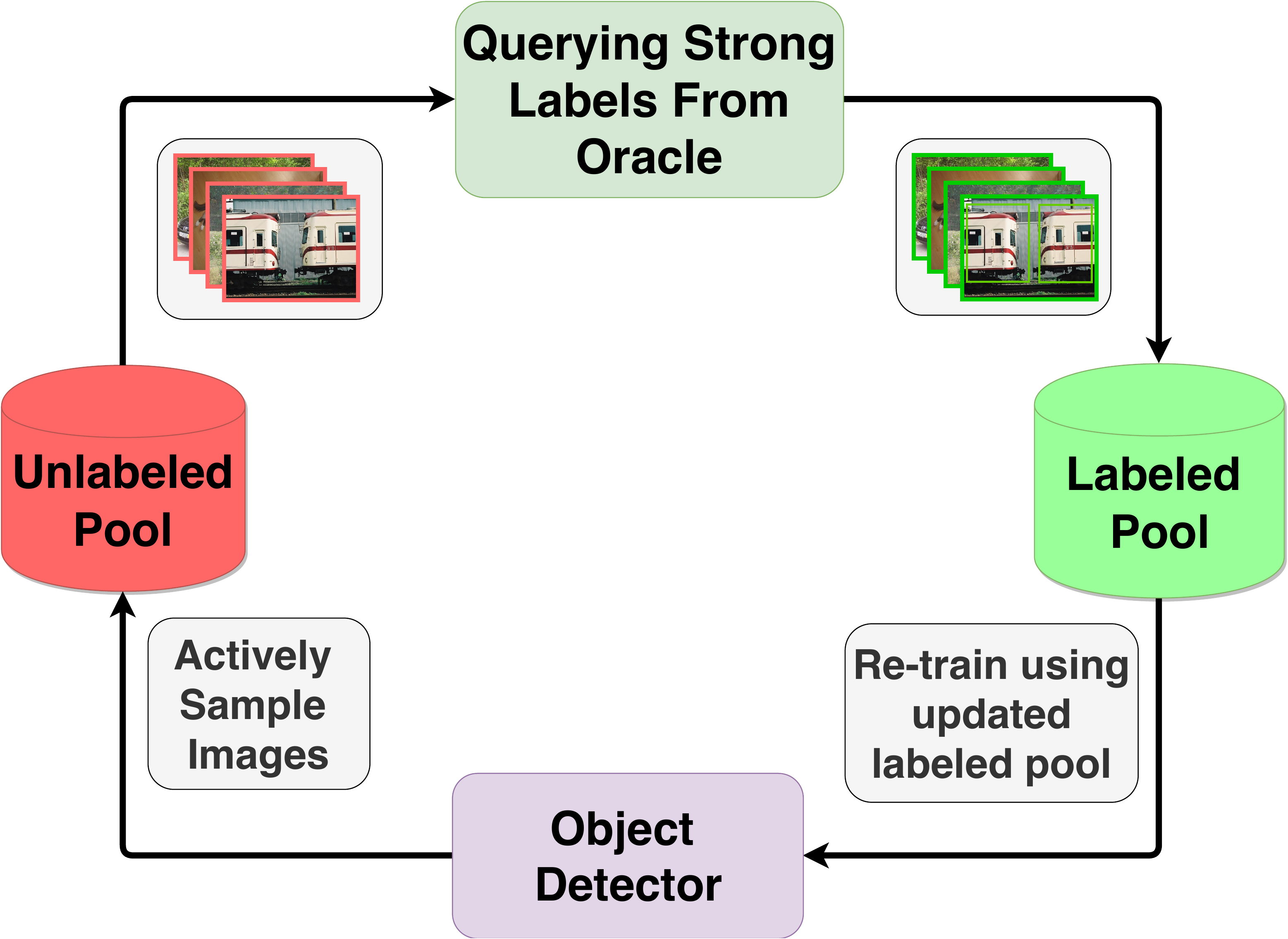}}&
\bmvaHangBox{\includegraphics[width=5.5cm, height=4cm]{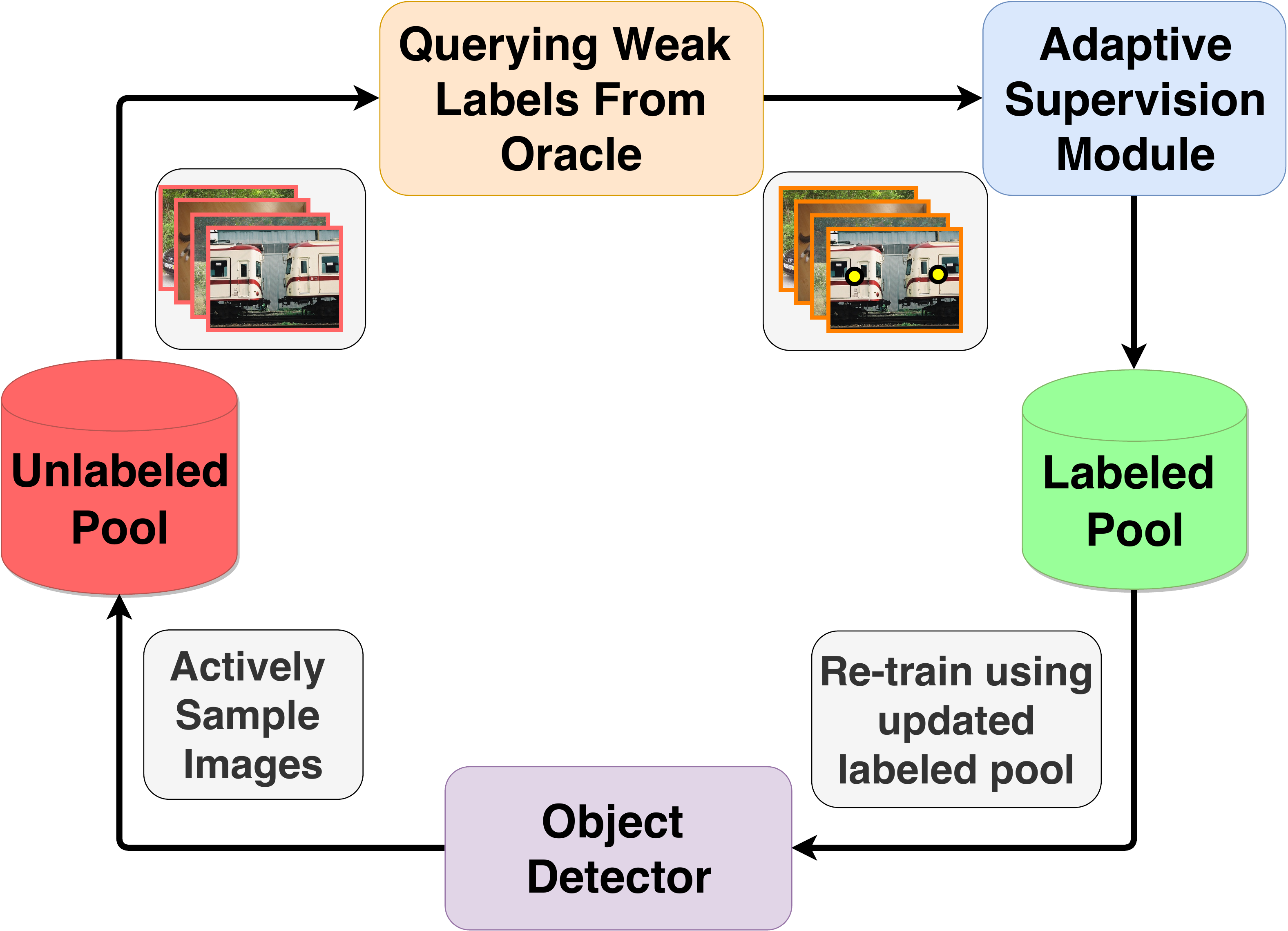}}\\
(a) Standard PBAL framework &(b) Proposed Framework
\end{tabular}
\caption{(a) Standard pool-based active learning (PBAL) framework; (b) Proposed framework which interleaves weak supervision in the active learning process. Our framework includes an adaptive supervision module which allows switching to a stronger form of supervision as required when training the model.}
\vspace{-12pt}
\label{fig:teaser}
\end{figure}

Learning with weak supervision has grown significantly in importance over the last few years \cite{wsol1,wsol2,wsol3,wsol4,wsol5,wsol6/corr/abs-1709-01829,click,dialog,extreme,Papadopoulos2016WeDN}. Achieving desired generalization performance with a lower labeling budget has been achieved using image-level labels \cite{wsol1,wsol2,wsol3,wsol4,wsol5,wsol6/corr/abs-1709-01829}, object center clicks \cite{click} and answering yes/no questions \cite{Papadopoulos2016WeDN}. On the other hand, active learning is a set of methods where the model systematically queries labels for the most informative subset of a given dataset. There has been no effort so far, to the best of our knowledge, that leverages weak supervision for better performance in active learning. While weak supervision focuses on learning with cheaper labeling methods, active learning focuses on reducing the number of samples required to label, with full supervision. These two classes of methods differ in their approach of reducing annotation costs. We propose that a combination of weak supervision and active learning can result in greater savings in annotation costs since both the label quality and the size of labeled data can be optimized. In this work, we propose an adaptive supervision framework for active learning and show its effectiveness in training object detectors. We use the standard pool based active learning approach, but instead of querying strong bounding box annotations (which are time consuming), we query a weaker form of annotation first and only query bounding box labels when required. We propose variants in how weak and strong supervision can be interleaved to show the flexibility of the proposed methodology. An overview of our framework is shown in Figure \ref{fig:teaser}. We validate the proposed methodology on standard datasets such as PASCAL VOC 2007 and VOC 2012, as well as in a real-world setting, agriculture, where labeling expertise is expensive, and the proposed methodology can provide significant savings in labeling budgets. 

\vspace{-12pt}
\section{Related Work}
\label{sec:rel}
\vspace{-5pt}
Previous work on reducing labeling efforts for training object detection methods can be broadly divided into two categories: Weak Supervision and Active Learning. Weakly supervised learning methods focus on reducing the labeling effort for each label, but however result in lower performance due to the imprecise supervision. Active learning methods focus on selecting appropriate image data for querying for labels in an iterative manner, but require fully supervised labels in each iteration.

\smallskip
\noindent
\textbf{Weak Supervision.} Image-level labels, i.e. the class names of the objects present in the image, are the most common form of weak supervision used in object detection. There have been several efforts on Weakly Supervised Object Localization (WSOL) \cite{wsol1,wsol2,wsol3,wsol4,wsol5,wsol6/corr/abs-1709-01829}, in which the task is to localize objects in an image given only image-level labels. However, models trained on image level labels typically do not reach the performance level of their fully supervised counterparts. Recently, alternative methods for annotating objects such as center-clicking \cite{click}, clicking on the object extremes \cite{extreme} and bounding box verification \cite{dialog,Papadopoulos2016WeDN} have been proposed, which show promising savings in annotation time. However, to the best of our knowledge, weak supervision methods have by far not integrated active learning into their training methodology. 

\smallskip
\noindent
\textbf{Active Learning.} This is a class of techniques used to pick the most beneficial samples to train a model (please see \cite{Settles10activelearning} for a detailed survey). Active learning has been shown to be very effective in image classification \cite{Sener2018ActiveLF,Gal2017DeepBA,Wang_gupta,al_gaussian,adaptive_al}. In a deep network setting, there have been limited efforts however in active learning for object detection \cite{Kao2018LocalizationAwareAL,brust,VinayNamboodiri2018}. Active selection metrics such as localization uncertainty \cite{Kao2018LocalizationAwareAL}, margin sampling based on convolutional layers \cite{VinayNamboodiri2018} and 1-vs-2 margin sampling \cite{brust} have been proposed. However, these methods directly query for full supervision during active learning. 

\smallskip
\noindent
In this work, we leverage the advantages of both these categories of methods by introducing an adaptive supervision framework for active learning in object detection. Our framework allows switching between weak and strong supervision to obtain significant savings in annotation cost when compared to earlier works. 

\vspace{-17pt}
\section{Methodology}
\label{sec_method}
\vspace{-6pt}
We first present an overview of the proposed framework, before explaining each component in detail. For the rest of the paper, we interchangeably use the terms weak supervision, weak labels and weak annotations. 

\vspace{-15pt}
\subsection{Overview}
\vspace{-5pt}
Figure \ref{fig:teaser}a shows the standard pool-based active learning (PBAL) setting for object detection, in which a batch of informative images is queried for bounding box annotations every episode, using which the object detector is updated. In the proposed method, instead of directly querying for time-consuming bounding box annotations, we first query for just weak labels and generate pseudo labels to train the model. Secondly, we introduce an adaptive supervision module to allow switching to strong supervision when required. We introduce two variants of supervision switching, namely \textit{hard switch} and \textit{soft switch}. A \textit{hard switch}, also called \textit{inter-episode switch}, causes the model to permanently switch to a stronger form of supervision at a certain stage of the training process, and after the switch, our framework reduces to a standard PBAL setting (Figure \ref{fig:teaser}a). In contrast, a \textit{soft switch}, also called \textit{intra-episode switch}, allows the model to query both forms of supervision in each round of active learning all through the training process. Based on a switching criterion, in a given batch of actively selected images, the model asks for weak supervision on some images and asks for strong supervision on the other images. More details are provided in Section \ref{sec:inc}. In brief, we show in our experiments that our adaptive supervision module results in substantial savings in annotation time. 

\vspace{-8pt}
\subsection{Active Learning Setup}
\vspace{-5pt}
\label{sec:al}
To begin with, we consider a deep object detection model $\mathcal{M}$ (e.g. Faster R-CNN \cite{faster}) and a dataset $\mathcal{D}$ which is initially unlabeled. Our aim is to maximize the model performance under a given labeling budget $\mathcal{B}$. We assume, like any other active learning setup, that an initial (randomly chosen) subset of $\mathcal{D}$ is queried for strong labels and a labeled pool of samples, $\mathcal{L}$, is generated. The remaining images form the unlabeled pool $\mathcal{U}$. We also consider a weakly labeled pool $\mathcal{W}$ which is initially empty. As a common practice in active learning, we begin with training our model $\mathcal{M}$ on the initial labeled pool. 

\begin{figure}[t]
\begin{tabular}{cc}
\hspace{+25pt}
\bmvaHangBox{\includegraphics[width=5cm, height=3.5cm]{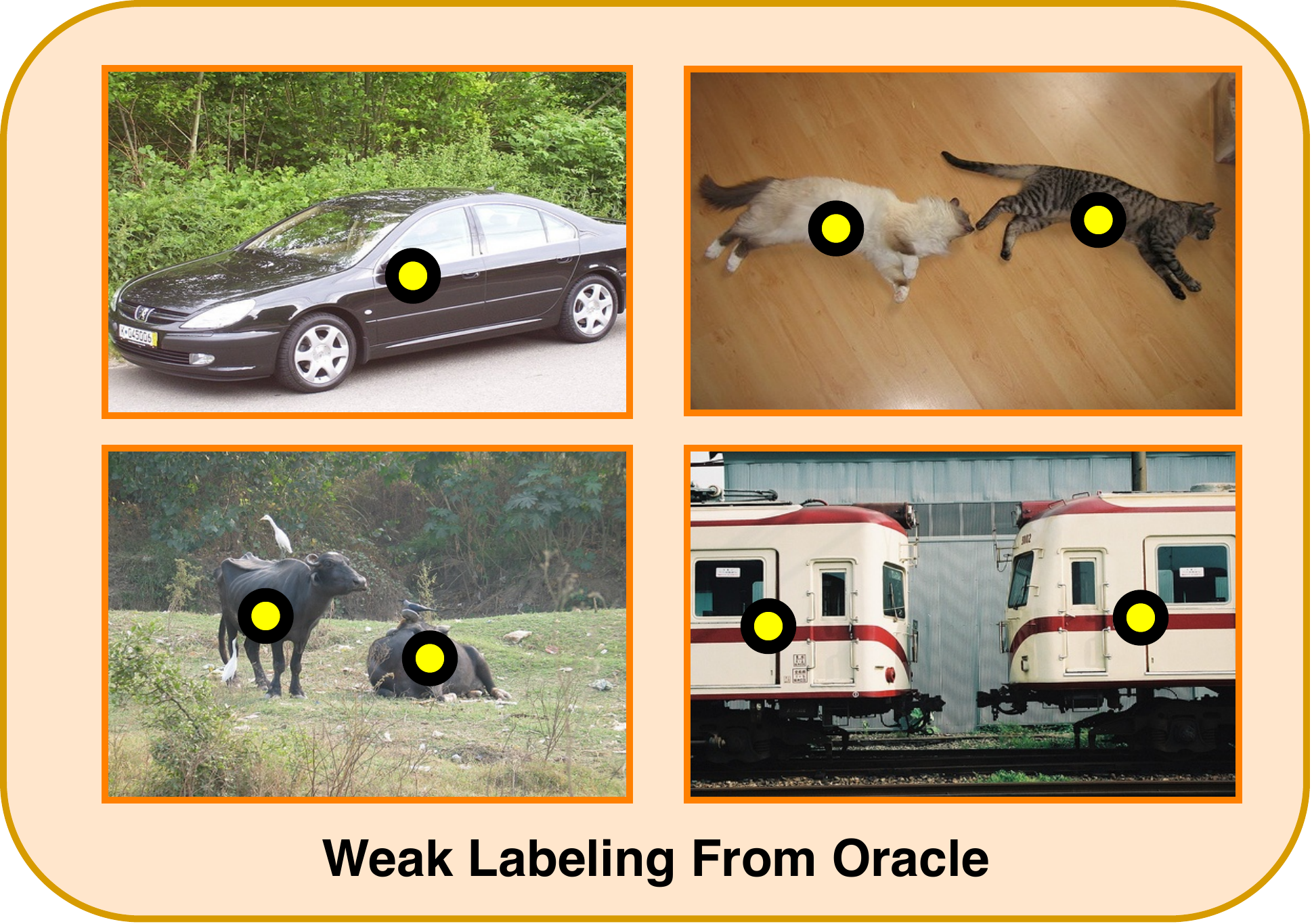}}& 
\bmvaHangBox{\includegraphics[width=5cm, height=3.5cm]{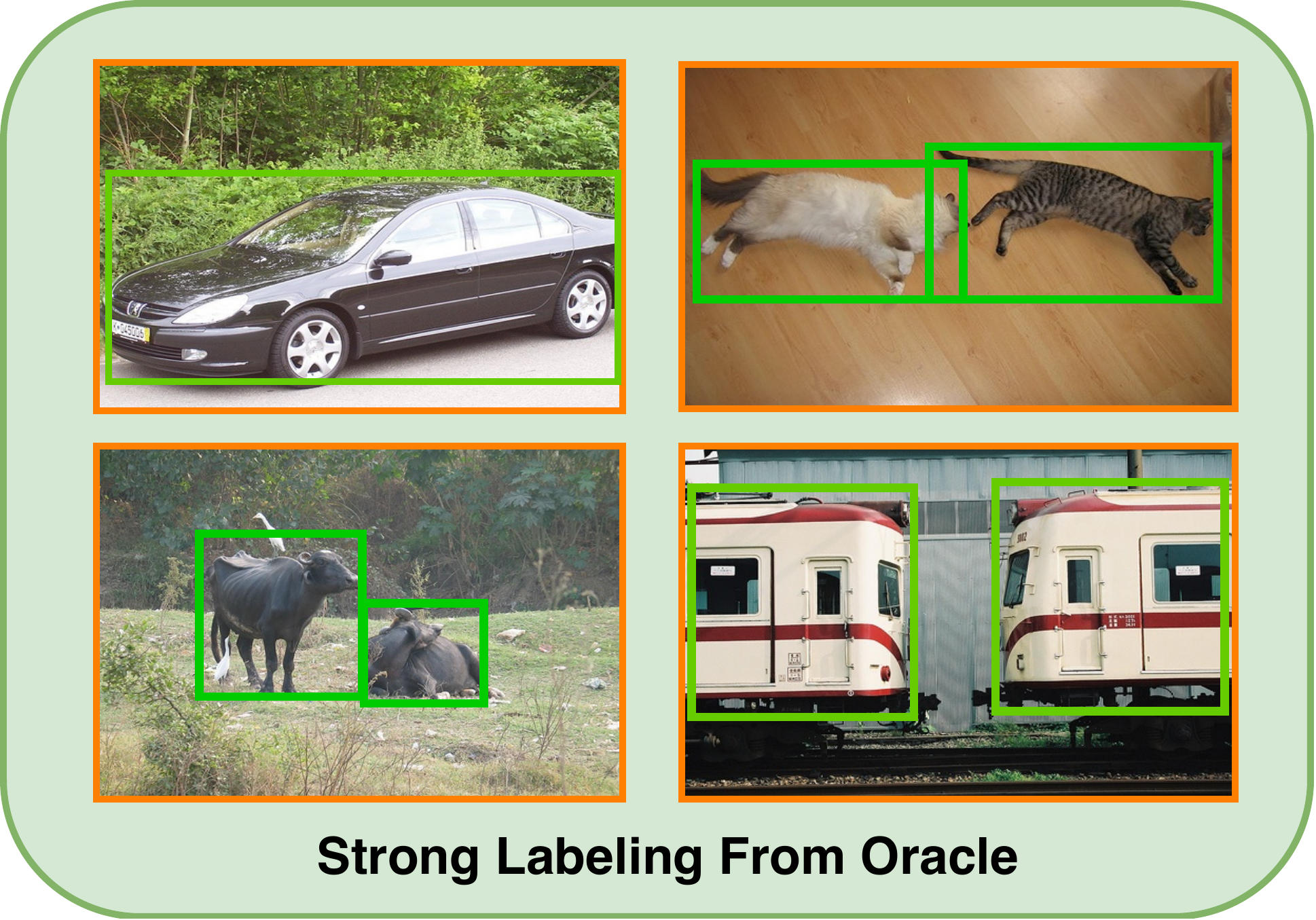}} \\
\hspace{+25pt}
(a)& 
(b)
\end{tabular}
\caption{Illustration of: (a) Weak supervision using center clicks, (b) Strong supervision using bounding box annotations and (c) Soft Switching Mechanism.}
\vspace{-15pt}
\label{fig:label}
\end{figure}


The choice of query technique is a key design decision in any active learning method. We study the use of multiple standard query techniques in this work, and show that given any of these querying techniques, our framework can achieve annotation savings when compared to the standard fully-supervised PBAL setting.


\vspace{-8pt}
\subsection{Labeling Techniques}\label{sec:labeling}
\vspace{-5pt}
In our framework, the oracle (e.g. a human annotator) can be queried for two types of annotations: 
\vspace{-5pt}
\begin{itemize}
\setlength\itemsep{0em}
    \item \textbf{Strong Labels:} Strong labeling involves drawing tight bounding boxes around objects in an image, and is the conventional form of labeling used for object detection datasets. Since the annotation times of the datasets we used were unavailable, we use the statistics of ImageNet \cite{Su2012CrowdsourcingAF} for consistency, as the difficulty and quality of annotations of PASCAL VOC and ImageNet are quite similar. Su \etal \cite{Su2012CrowdsourcingAF} and Papadopoulos \etal \cite{click} report the following median annotation times on ImageNet: 25.5s for drawing one box, 9.0s for verifying its quality and 7.8s for checking whether there are other objects of the same class yet to be annotated.  Hence, we take 34.5s (25.5s + 9.0s) to be the median time taken to draw an accurate bounding box around an object and additionally add 7.8s for every image annotated.
    
    \item \textbf{Weak Labels: } In our method, we use the recently proposed center-clicking \cite{click} as our weak labeling method. For each object in a given image, the annotator clicks approximately on the center of the imaginary bounding box that encloses the object. Papadopoulos \etal \cite{click} report that the maximum median time to click on an object's center is 3.0s.
\end{itemize}
\vspace{-5pt}
Figure \ref{fig:label} illustrates both the above mentioned labeling techniques used in our framework.

\begin{figure}[b]
\vspace{-12pt}
\begin{tabular}{cc}
\bmvaHangBox{\includegraphics[width=7.5cm, height=2.8cm]{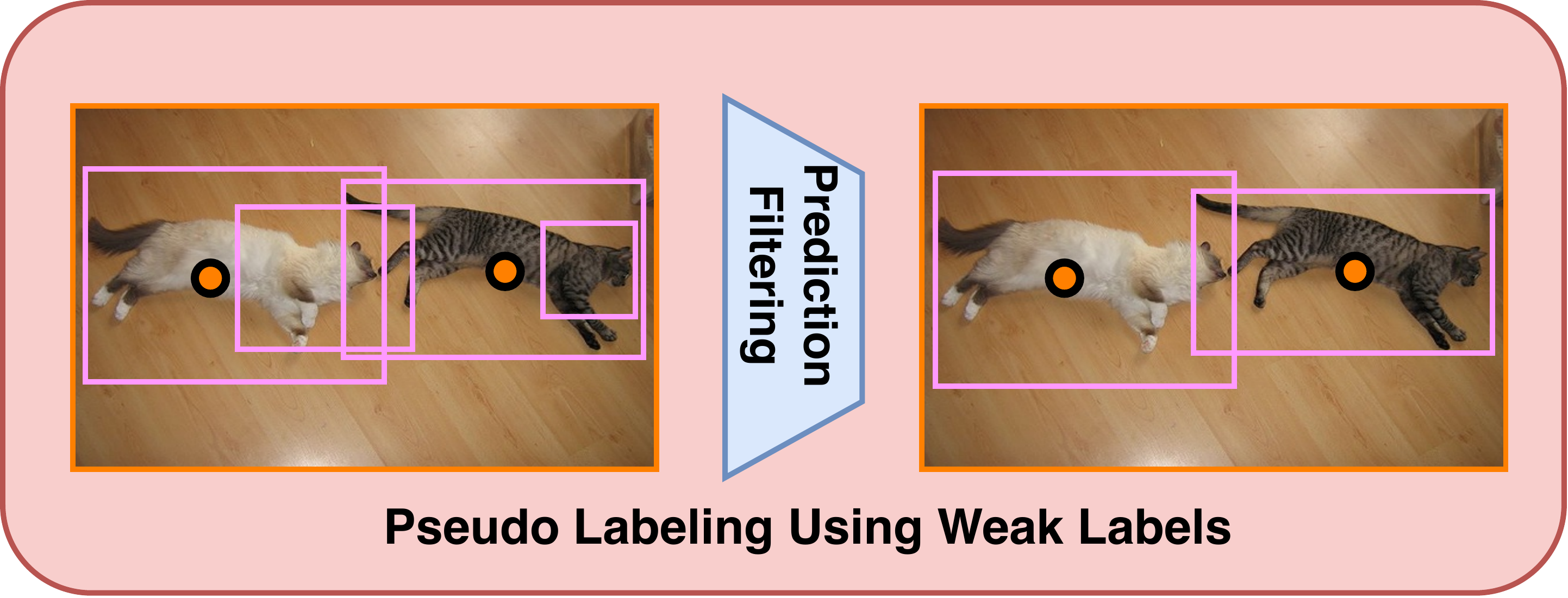}}&
\bmvaHangBox{\includegraphics[width=4.5cm, height=2.8cm]{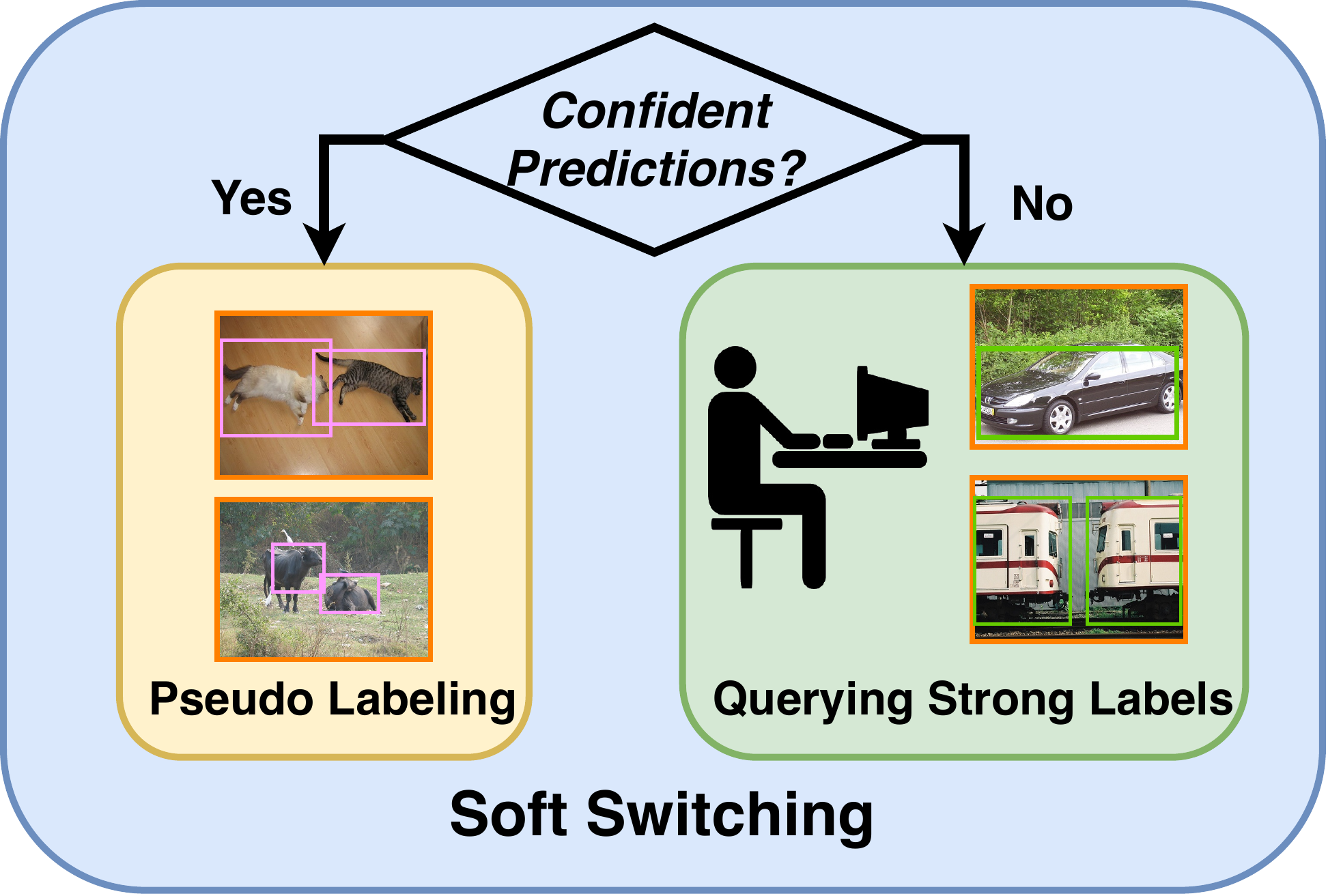}}\\
(a)&(b)
\end{tabular}
\caption{We illustrate (a) pseudo labeling using center clicks and (b) the soft switching mechanism used in the adaptive supervision module.}
\vspace{-5pt}
\label{fig:misc}
\end{figure}

\vspace{-12pt}
\subsection{Adaptive Supervision}
\label{sec:inc}
\vspace{-5pt}
We use the adaptive supervision module which helps in deciding when its time to make a switch from weak to strong supervision. 
A stronger supervision method takes up more annotation time but provides greater information to the model. We propose two variants of supervision switching, which are at different levels of granularity: 
\begin{enumerate}
\setlength\itemsep{0em}
    \item \textbf{Hard (Inter-Episode) Switch:} As the name suggests, in this method, we define a switching criterion at the end of a given episode based on the change in model's performance on the validation set. Let $d_n$ be the difference between mAP of the model in episode $n$ and $n-1$, and $d_{max}$ be the maximum difference of mAP between any two consecutive episodes until episode $n$ and $\gamma \in [0,1]$ be a suitably chosen threshold value. The criterion can then be written as follows:
    \vspace{-3pt}
   \begin{equation}
  S_{hard}(n) = \left\{
     \begin{array}{@{}l@{\thickspace}l}
       \text{1}  & \hspace{0.5cm} \text{if } \frac{d_n}{d_{max}} \leq \gamma \\
       \text{0}  & \hspace{0.5cm} \text{otherwise}  \\
     \end{array}
   \right.
   \label{eq:1}
\end{equation}
    When the above condition evaluates to 1, we perform a \textit{hard switch} (and hence the name) to strong supervision i.e., our model would query only strong bounding box annotations in later episodes of active learning thus reducing to a standard PBAL setup then on (Figure \ref{fig:teaser}a).
    \item \textbf{Soft (Intra-Episode) Switch:} In each episode of active learning, we use the obtained weak labels for the actively selected batch to \textit{pseudo-label} these selected images with a bounding box. Pseudo-labeling the images is a simple low-cost step as described in Section \ref{sec:pseudo}. For each image, we obtain a confidence score $c$ which is the mean probability score obtained for each predicted object. Given the confidence score $c_i$ for a selected image $i$ and a suitably chosen threshold $\delta \in [0,1]$, we perform the soft switch when the following condition evaluates to 1:
    \vspace{-7pt}
    \begin{equation}
  S_{soft}(i) = \left\{
     \begin{array}{@{}l@{\thickspace}l}
       \text{1}  & \hspace{0.5cm} \text{if } c_i < \delta  \\
       \text{0}  & \hspace{0.5cm} \text{otherwise}  \\
     \end{array}
   \right.
\end{equation}
In other words, we query an image for strong supervision if the model's average confidence on its object predictions is below a threshold $\delta$. Otherwise, we pseudo label the image using its current predictions. This intuitively makes sense because we query for strong labels only when the model is very unsure of its current bounding box predictions, else manage with the weak annotations. We note that this switch is carried out episode-wise, and each new episode starts afresh with seeking weak labels again for images with a reasonably high confidence (and hence, the name \textit{soft switch}).
\end{enumerate}
\vspace{-12pt}
\subsection{Pseudo Labeling using Weak Labels}
\label{sec:pseudo}
\vspace{-5pt}
We use a low-cost pseudo-labeling approach to train object detectors with weakly labeled data. We do not explicitly consider complex training methods for learning with weak supervision \cite{click,mil1,mil2} to avoid introducing significant computational overhead in the training methodology and to keep our approach as model-agnostic as possible. In this approach, we first use the trained model, $\mathcal{M}$, to predict bounding boxes (which may be imprecise) for all possible classes on the weakly labeled images. We then use the weak labels provided by the oracle to filter and choose the best possible bounding box for each object as follows. 


In a given weakly labeled image, each center-click (our weak annotation) location corresponds to an object. For every click location, we pseudo-label that object with a bounding box with center closest to the click location. The object is classified as the class with the highest probability for the chosen bounding box. Computationally, this method involves a forward pass through the network for each image followed by computation of pairwise distances (2 dimensions) between the click locations and centers of predicted bounding boxes. Figure \ref{fig:misc}a illustrates our pseudo-labeling strategy. We finally use labeled data (from $\mathcal{L}$) and pseudo-labeled data (from $\mathcal{W}$) to retrain our object detection model in an end-to-end manner. Our overall methodology is summarized below in Algorithm \ref{algo1}.

\begin{algorithm}[H]
\label{algo1}
\DontPrintSemicolon
\SetAlgoLined
\SetKwInOut{Input}{Input}
\SetKwInOut{Output}{Output}
\SetKwFunction{ActiveSampling}{ActiveSampling}
\SetKwFunction{QueryWeakAnnotations}{QueryWeakAnnotations}
\SetKwFunction{QueryAnnotations}{QueryAnnotations}
\SetKwFunction{PseudoLabels}{PseudoLabels}
\Input{Unlabeled pool $\mathcal{U}$, Labeled pool $\mathcal{L}$, Weak labeled pool $\mathcal{W}$, Model $\mathcal{M}$, episode num $n$, sample size $b$, soft switch threshold $\delta$, hard switch threshold $\gamma$, query function \ActiveSampling}
\Output{Updated model $\mathcal{M}$}
\BlankLine

// Actively sample $b$ valuable images \;
$S$ = \ActiveSampling(from \{$\mathcal{U \cup \mathcal{W}}$\}, sample size = $b$) \;
// Query weak annotations on $S$ \;
$W_S$ = \QueryWeakAnnotations($S$) \;
// Obtain pseudo labels for $S$ using $W_S$, as described in Sec \ref{sec:pseudo} \;
$P_S$ = \PseudoLabels(model = $\mathcal{M}$, sample = $S$, weak supervision = $W_S$) \;
\uIf{soft switch}{
$S_{high} := \{i : i \in S\ \ni \text{confidence}(P^i_S) > \delta \}$ \;
$S_{low} := \{i : i \in S\ \ni \text{confidence}(P^i_S) \leq \delta \}$ \;
// Use pseudo labels for $S_{high}$\;
$S_{high}^{Pseudo} := \{P^i_S : i \in S_{high}\}$\;

// Query strong annotations on $S_{low}$  \;
$S_{low}^{Strong}$ := \QueryAnnotations($S_{low}$) \;
$\mathcal{L} \leftarrow \mathcal{L} \cup S_{low}^{Strong}$; $\mathcal{W} \leftarrow \mathcal{W} \cup S_{high}^{Pseudo}$ \;
}
\uElseIf{hard switch}{
$d$ := difference in $mAP$ between last two episodes\; 
$d_{max}$ := maximum difference in $mAP$ between episodes so far\;
\uIf{$\frac{d}{d_{max}} \leq \gamma$}{
      Use fully supervised pool-based active learning from next episode \;
}
$\mathcal{W} \leftarrow \mathcal{W} \cup P_S$\;
}
Train model $\mathcal{M}$ on \{$\mathcal{L} \cup \mathcal{W}$\} \;
return $\mathcal{M}$

\caption{Adaptive Supervision for Active Learning}
\end{algorithm}

\vspace{-10pt}
\section{Experiments and Results}
\vspace{-5pt}
\subsection{Implementation Details}
\vspace{-5pt}
\noindent
\textbf{Active Sampling Techniques.} The choice of query technique is a key design decision in any active learning method. Considering our framework is independent of the query method, We study the following query techniques to actively sample images: \textbf{(i) Max-Margin:} For a predicted bounding box, margin is calculated as the difference between the first and the second highest class probabilities. For each image, margin is chosen to be the summation of margins across all the predicted bounding boxes in the image, as in Brust \etal \cite{al-deep-object-baseline}. \textbf{(ii) Avg-Entropy:} Samples with high entropy in the probability distribution of the predictions are selected, as in Roy \etal \cite{VinayNamboodiri2018}. \textbf{(iii) Least Confident:} Confidence for an image is calculated as the highest bounding box probability in that image. Images with least confidence are selected. This criterion is taken from the minmax method specified in \cite{VinayNamboodiri2018}. 

\smallskip
\noindent
\textbf{Evaluation Metrics:} In our experiments, we measure the annotation effort required to reach a certain level of test performance. Annotation effort is measured in terms of time taken to annotate images through the active learning cycle. As discussed in \ref{sec:labeling}, to have a consistent measure, we follow the previous work on click supervision\cite{click} and utilize the median annotation times reported on ImageNet \cite{Su2012CrowdsourcingAF} to compute time taken for bounding box annotations and weak annotations. In our experiments, we use object center clicks as the chosen form of weak supervision. To get weak supervision for our datasets, we obtain the centers of the ground truth bounding boxes and perturb the center location by a small zero mean Gaussian random noise for robustness. Given an image $I$ with $b_I$ objects, we hence calculate annotation time (in seconds) as:
\vspace{-5pt}
\begin{equation}
\label{eq:annotation-time-baselines}
Time(I) = 
\begin{cases*}
  7.8 + 34.5 \times b_I & for bounding box annotations \\
  7.8 + 3 \times b_I & for center click annotations \\
\end{cases*}
\vspace{-3pt}
\end{equation}   
We use mean Average Precision (mAP) to evaluate the performance of the detection itself.

\vspace{-8pt}
\subsection{Results on PASCAL VOC 2007}
\label{sec:voc7}    
\vspace{-5pt}
\noindent
\textbf{Setup.} We show results on PASCAL VOC 2007 \cite{pascal-voc-2007} with 20 object classes. We use the trainval set of 5011 images as our training set $\mathcal{D}$ and evaluate our model's performance on the test set of 4952 images. In all our experiments, we use Faster R-CNN \cite{faster} with ResNet-101 \cite{DBLP:journals/corr/HeZRS15} backbone as our object detection model, and extend PyTorch implementation by \cite{jjfaster2rcnn}. As in Section \ref{sec:al}, we follow the standard pool-based active learning (PBAL) setup. We choose 500 images (around 10\% of dataset) as the initial labeled pool $\mathcal{L}$ and train our model on it. 

\smallskip
\noindent
\textbf{Active Learning.} With the same initial model, we use different query techniques: max-margin sampling \cite{al-deep-object-baseline}, least-confident \cite{VinayNamboodiri2018} and  avg-entropy \cite{VinayNamboodiri2018}. For each query method, we implement the standard PBAL framework, our adaptive supervision framework with hard switching and soft switching. We do not do hard switching for the least confident sampling method because pseudo labeling the least confident samples using the model is counter-intuitive. We fix an annotation budget $\mathcal{B}$ of 35 hours. Until this budget is exhausted, we run multiple episodes of active learning. In any given episode, if we're querying for strong supervision, we query 250 images (around 5\% of the dataset size). If we're querying for weak supervision instead, we query 500 images (around 10\% of the dataset size). 

\smallskip
\noindent
\textbf{Adaptive Supervision.} While performing active learning with our adaptive supervision module, we set $\gamma = 0.3$ as our \textit{hard switch} threshold, i.e. we switch to strong supervision when the test mAP increase in the last episode is less than 30\% of the maximum test mAP increase in any previous episode. While using \textit{soft switch}, we set the probability threshold $\delta = 0.75$ i.e., if a model's average confidence on an actively sampled image is less than $0.75$, that image will be queried for strong labels.

\begin{figure}
\begin{tabular}{llc}
\bmvaHangBox{\includegraphics[width=3.8cm]{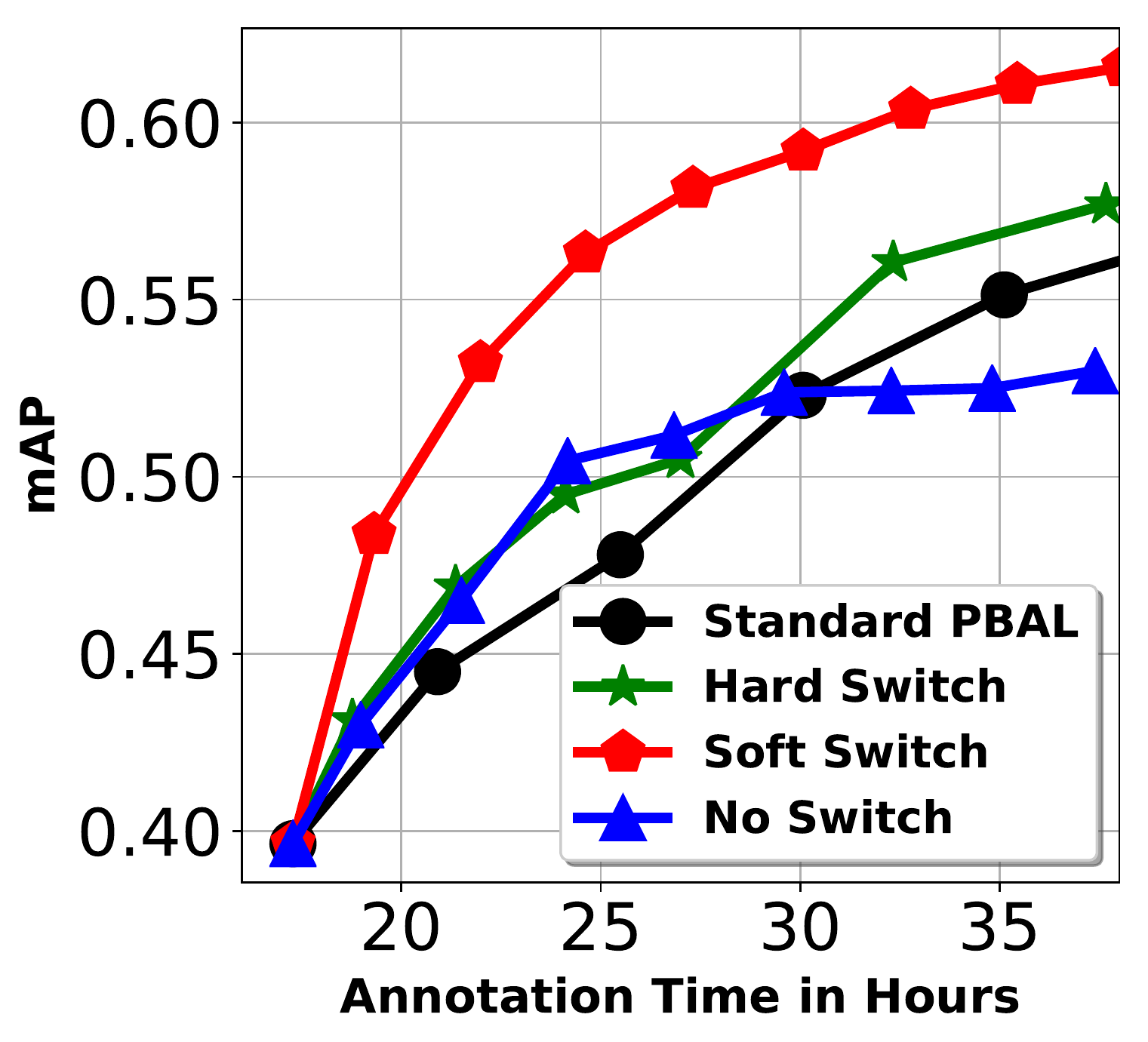}}&
\bmvaHangBox{\includegraphics[width=3.8cm]{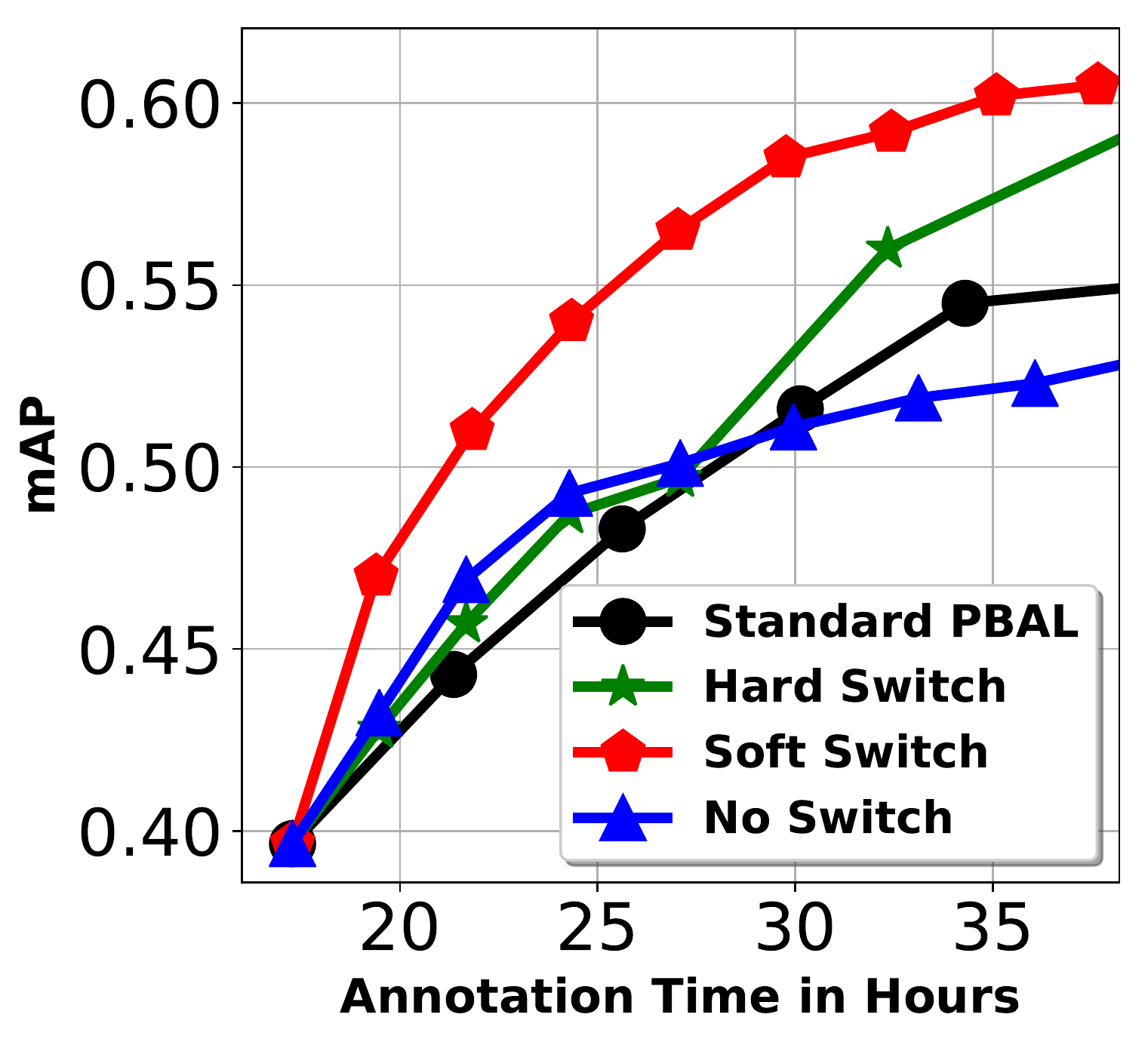}}&
\bmvaHangBox{\includegraphics[width=3.8cm]{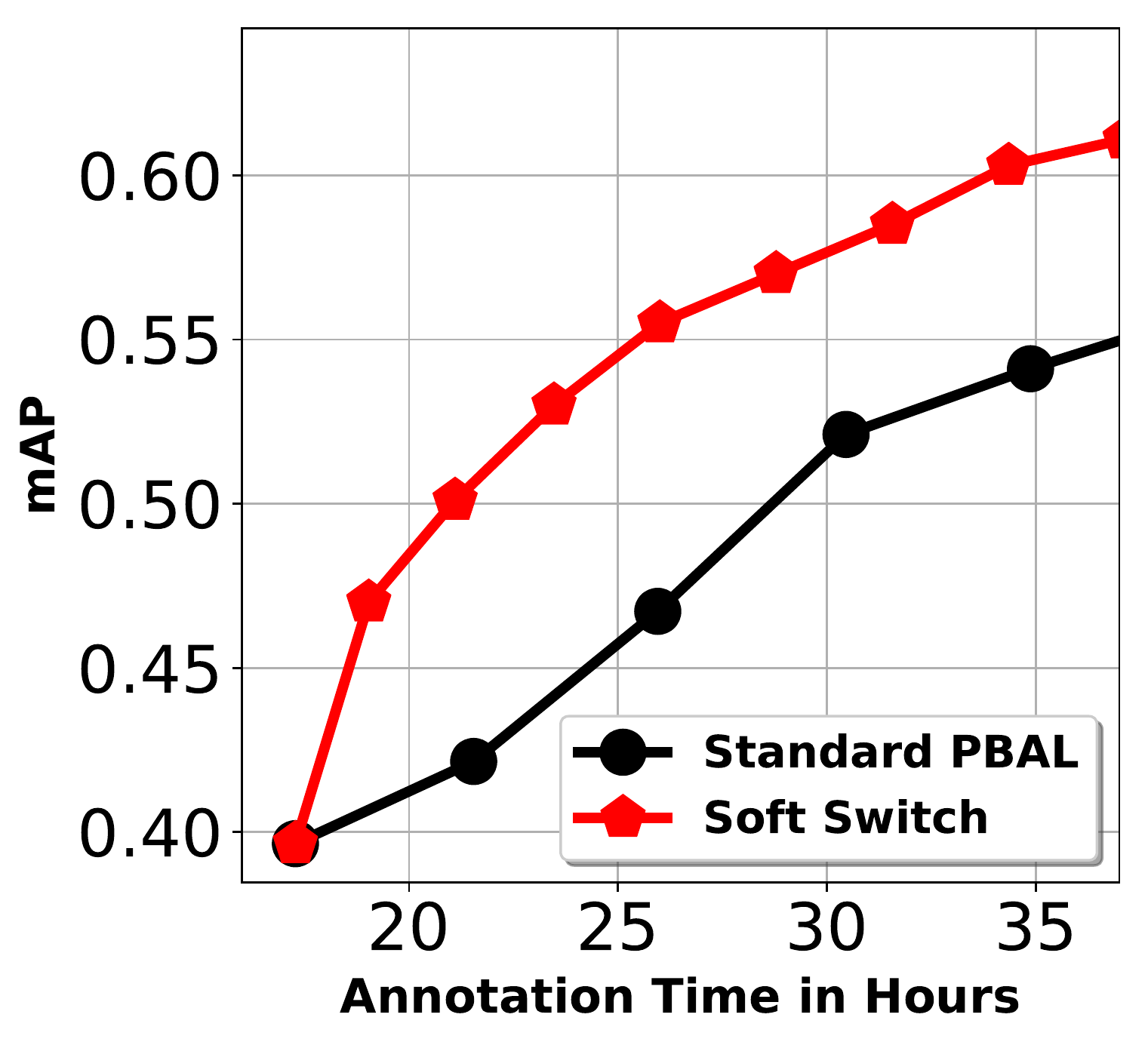}}\\
(a) Avg-Entropy &(b) Max-Margin & (c) Least-Confident
\end{tabular}
\caption{PASCAL VOC 2007: For each active query method, we show  performance of our adaptive supervision methods against standard PBAL framework (budget = 35 hours).}
\vspace{-10pt}
\label{fig:voc7_graphs}
\end{figure}

\begin{figure}
\begin{tabular}{c}
\hspace{+5pt}
\bmvaHangBox{\includegraphics[width=12cm, height=4cm]{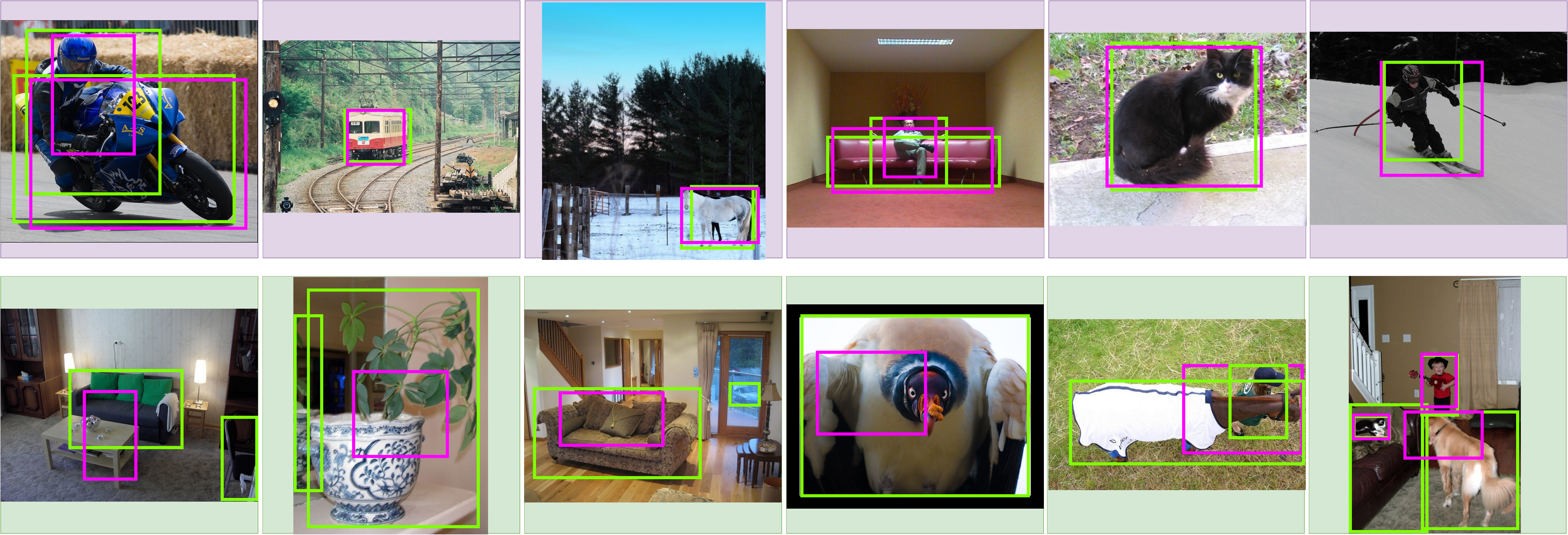}}\\
\end{tabular}
\caption{Soft switch mode on VOC 2007: (Top row) Images pseudo labeled using weak labels vs (Bottom row) images queried for strong labels. Boxes in pink were predicted while boxes in green denote ground truth.}
\vspace{-12pt}
\label{fig:pseudo-labeled-or-not}
\end{figure}

\smallskip
\noindent
\textbf{Evaluation.} Figure \ref{fig:voc7_graphs} shows the performance of various training methods for three different active sampling methods. In the figure, `Standard PBAL' represents the standard PBAL with strong supervision in every episode. The graphs corresponding to hard switch and soft switch represent our adaptive supervision methods. Finally, \textit{no switch} represents active learning using only weak supervision. It can be observed that our soft switch method significantly outperforms the standard PBAL method. For example, as seen in Figure \ref{fig:voc7_graphs}a, to achieve a test mAP close to 0.55, standard PBAL requires $\approx 35$ hours of annotation time whereas soft switch requires only 24.6 hours (30\% savings) and hard switch requires around 30 hours (14\% savings). Similarly the graphs for the other two metrics show that soft switch method achieves significant reduction in annotation efforts. A few qualitative results on VOC 2007 are shown in Figure \ref{fig:pseudo-labeled-or-not}, the top row images were pseudo labeled using just weak labels, the bottom row images were queried for strong labels. A significant difference in prediction quality can be observed between them. 
\vspace{-10pt}
\subsection{Results on PASCAL VOC 2012}
\vspace{-5pt}
\noindent
\textbf{Setup.} We perform similar experiments on PASCAL VOC 2012 \cite{pascal-voc-2012} which also has 20 object classes. We use the training set of 5717 images as our training set $\mathcal{D}$ and evaluate our model's performance on the validation set of 5823 images. The experimental setup is same as that of PASCAL VOC 2007, with an annotation budget $\mathcal{B}$ of 35 hours. We query 250 images  in a strong supervision episode and 500 images in a weak supervision episode. We use the same threshold values for adaptive supervision used for the experiments on PASCAL VOC 2007.

\begin{figure}
\begin{tabular}{llc}
\bmvaHangBox{\includegraphics[width=3.8cm]{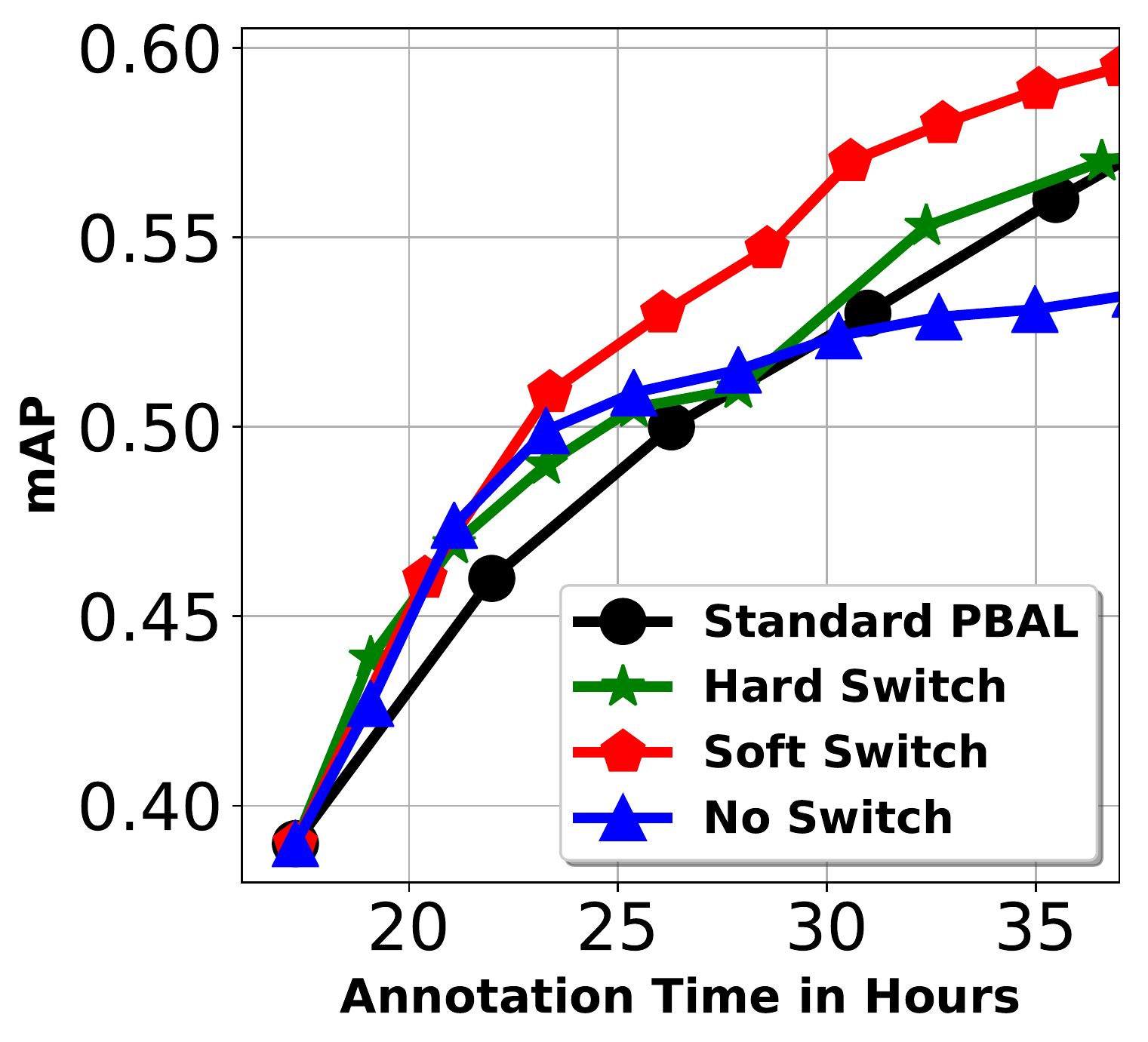}}&
\bmvaHangBox{\includegraphics[width=3.8cm]{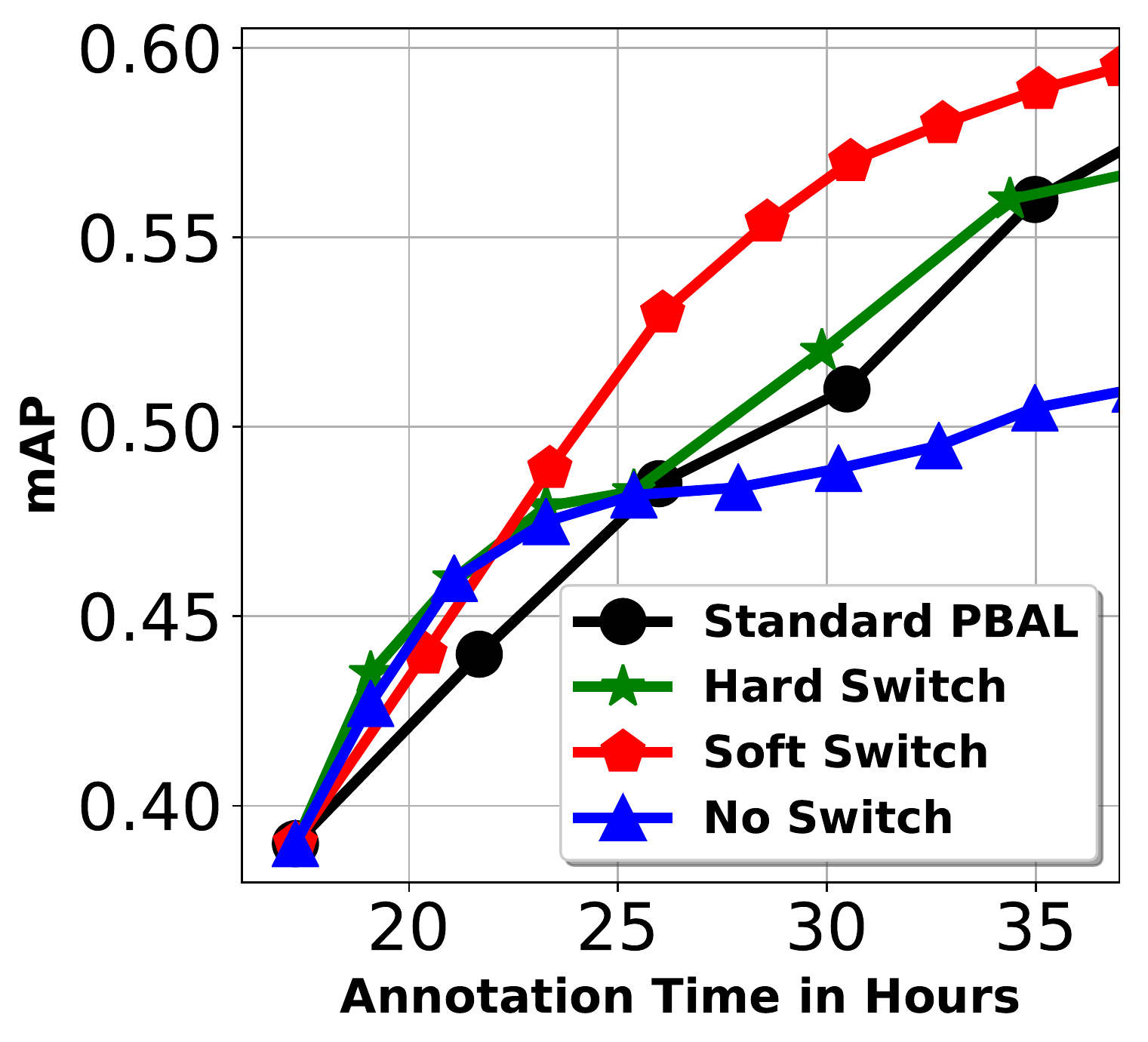}}&
\bmvaHangBox{\includegraphics[width=3.8cm]{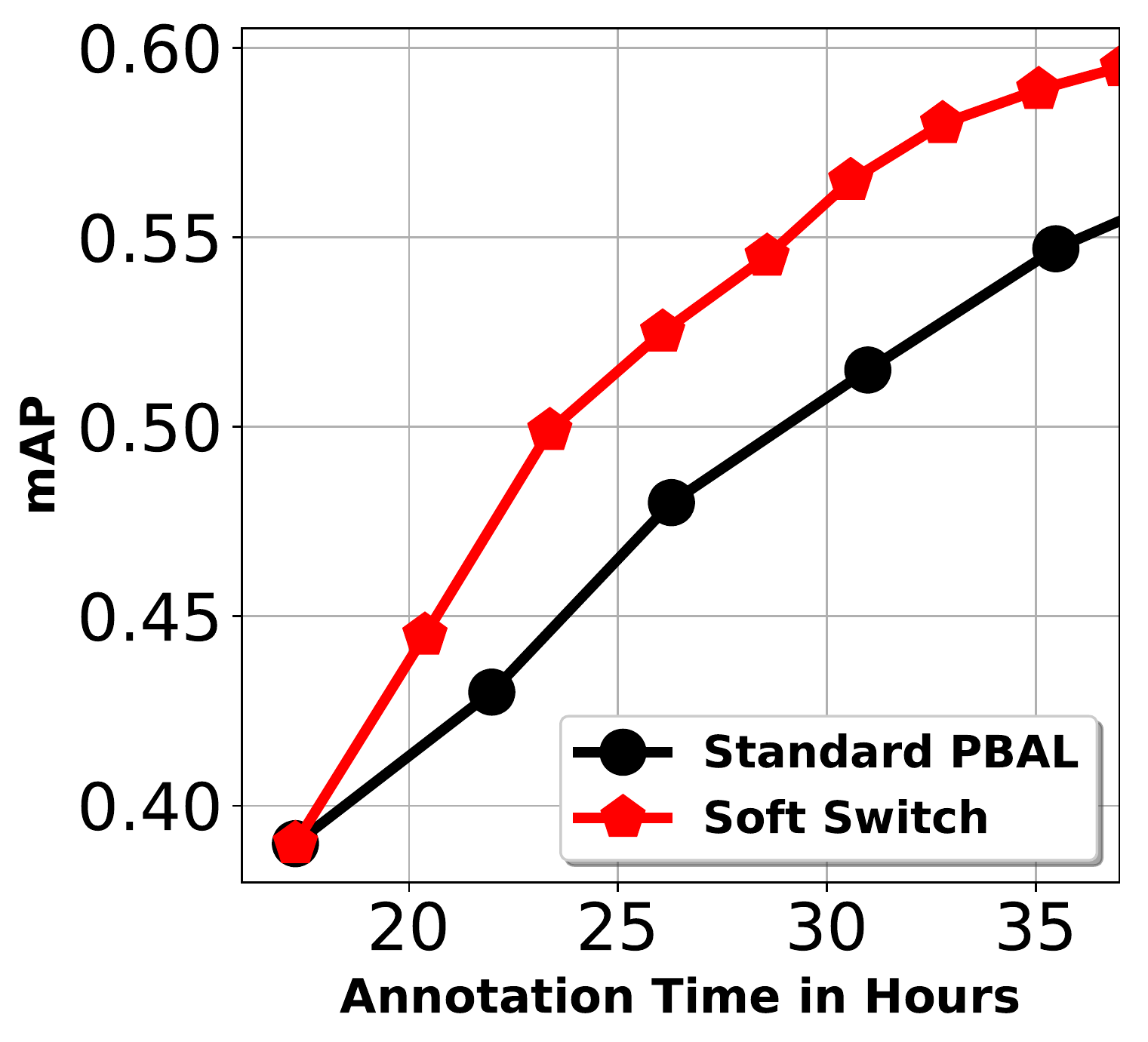}}\\

(a) Avg-Entropy &(b) Max-Margin & (c) Least-Confident
\end{tabular}
\caption{PASCAL VOC 2012: For each active query method, we show  performance of our adaptive supervision methods against standard PBAL framework (budget = 35 hours).}
\vspace{-11pt}
\label{fig:voc12_graphs}
\end{figure}

\smallskip
\noindent
\textbf{Evaluation.} We show the performance comparison of different supervision techniques for three different active sampling techniques in Figure \ref{fig:voc12_graphs}. Once again, soft switching outperforms all other compared methods. For example, in Figure \ref{fig:voc12_graphs}a, for achieving a test mAP close to 0.55, soft switching requires around 29 hours of annotation time (17.1\% savings) compared to 32.5 for hard switching (7.1\% savings) and 35 hours for the standard PBAL method. Also, hard switching slightly outperforms the standard PBAL method as seen in Figure \ref{fig:voc12_graphs}a and \ref{fig:voc12_graphs}b. Thus, hard switching can be used in a case where obtaining weak and strong supervision at the same time is not feasible.

\vspace{-8pt}
\subsection{Results on Wheat}
\vspace{-5pt}
\noindent
\textbf{Setup.} 
In addition to standard datasets, we show the effectiveness of our adaptive supervision framework on a real world agriculture dataset of wheat images. Obtaining expert level labels on agricultural datasets is generally expensive. In this case, we show that our framework can result in significant savings in labeling efforts. We use the Wheat dataset by Madec \etal. \cite{wheat}, which contains high definition images of wheat plants with objects of a single class: wheat head. To create a dataset suitable for training a deep object detection network, we preprocess the original $4000 \times 6000$ images as follows. We first downsample the images by a factor of 2 (to $2000 \times 3000$) using a bi-linear aggregation function. We then split these downsampled images into tiles of $500 \times 500$ images with no overlap (this has no impact on the study due to the nature of these images). We split the set of obtained 5663 images to use 4530 images (80\%) for training and 1133 images (20\%) for testing our methods. We choose 450 images (around 10\% of the dataset size) as the initial labeled pool $\mathcal{L}$ and train our model on it.


\noindent
\textbf{Active Learning and Adaptive Supervision.} Since this is a dataset with a highnumber of object instances in each image, we use an annotation budget $\mathcal{B}$ of 50 hours. Until this budget is exhausted, we run multiple episodes of active learning. In any given episode, if we're querying for strong supervision, we query 250 images. If we're querying for weak supervision instead, we query 500 images. While performing active learning with our adaptive supervision module, we set $\gamma = 0.3$ as our \textit{hard switch} threshold. While using \textit{soft switch}, we set the probability threshold $\delta = 0.85$.

\begin{figure}
\begin{tabular}{llc}
\bmvaHangBox{\includegraphics[width=3.8cm]{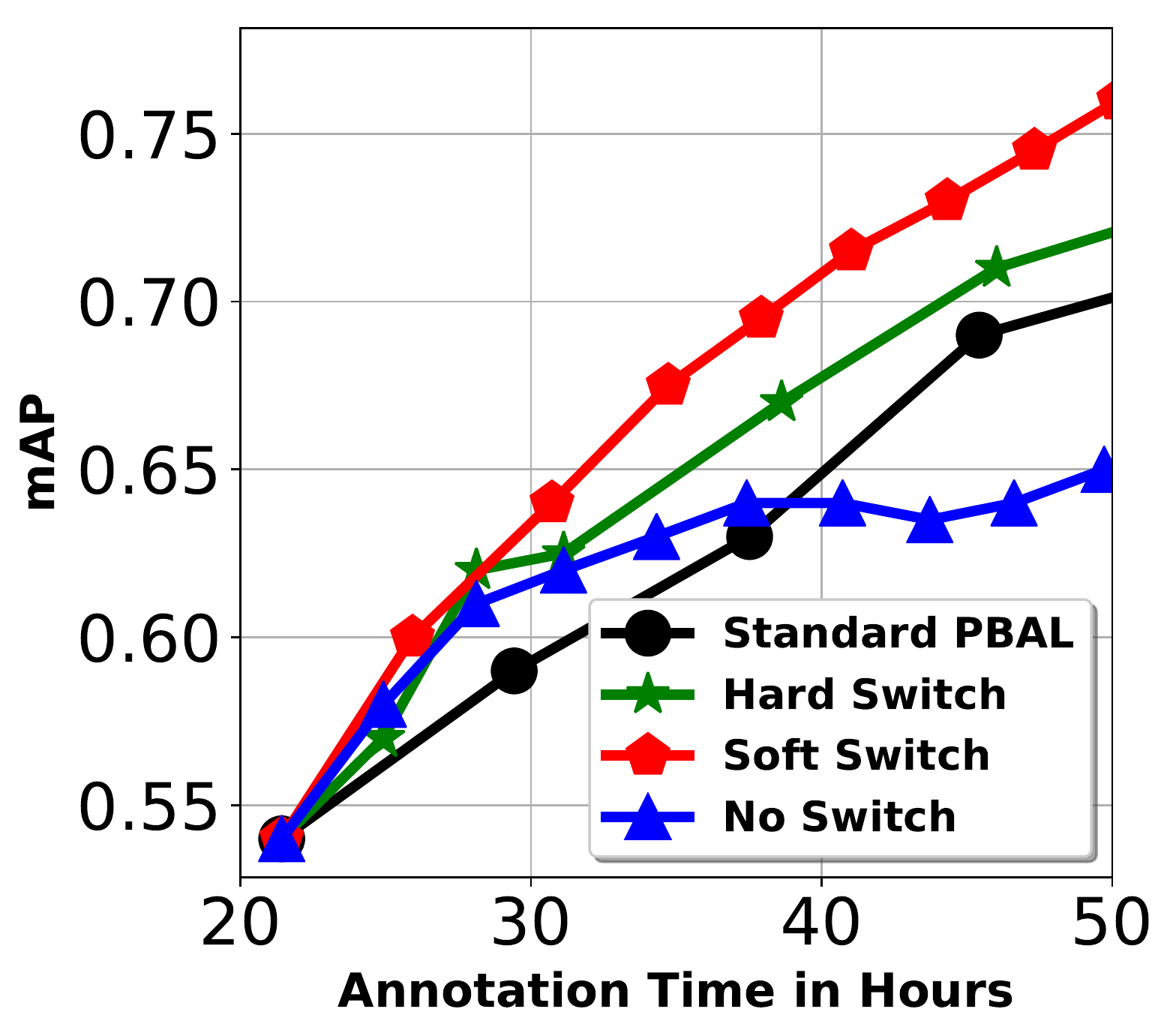}}&
\bmvaHangBox{\includegraphics[width=3.8cm]{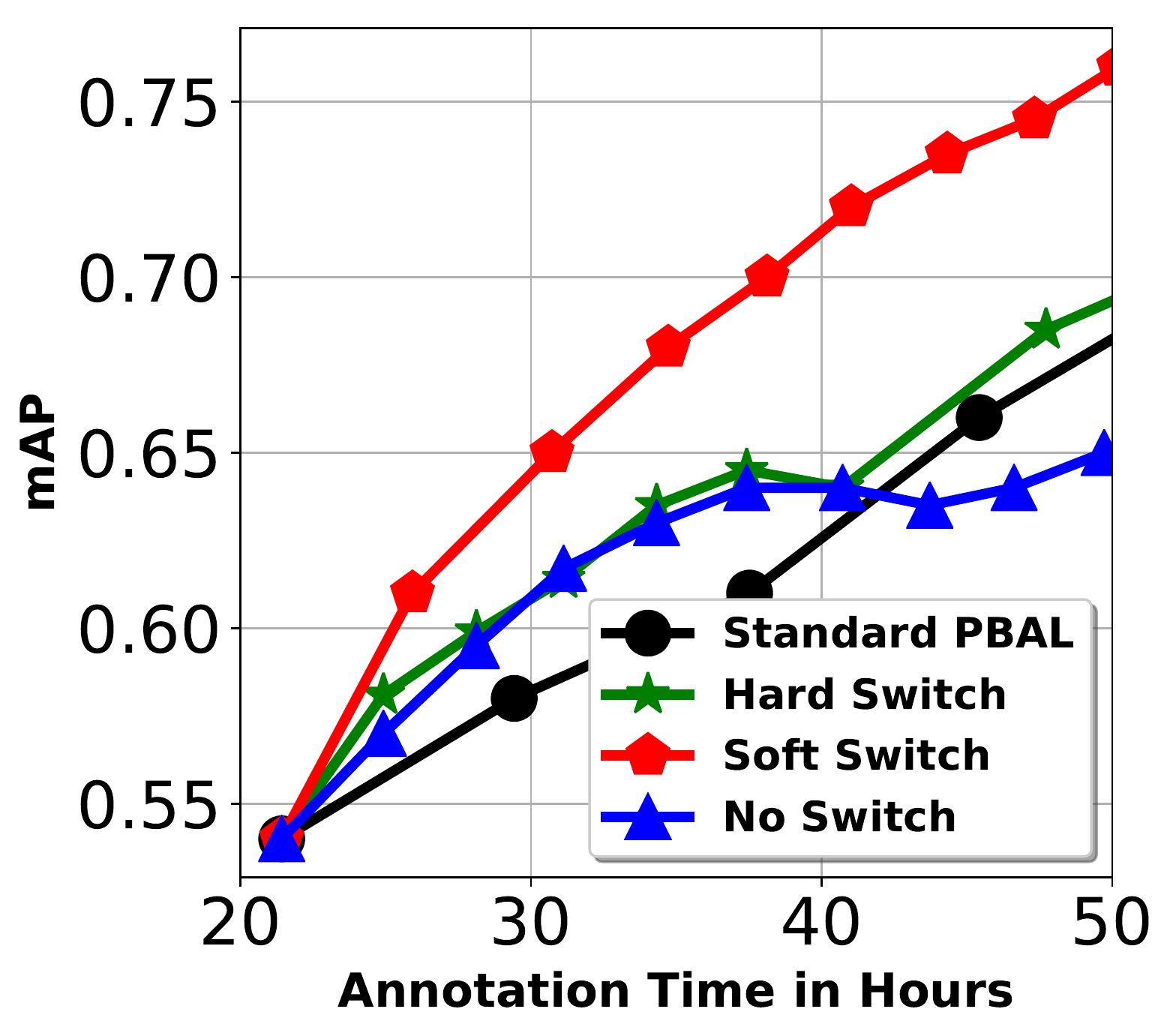}}&
\bmvaHangBox{\includegraphics[width=3.8cm]{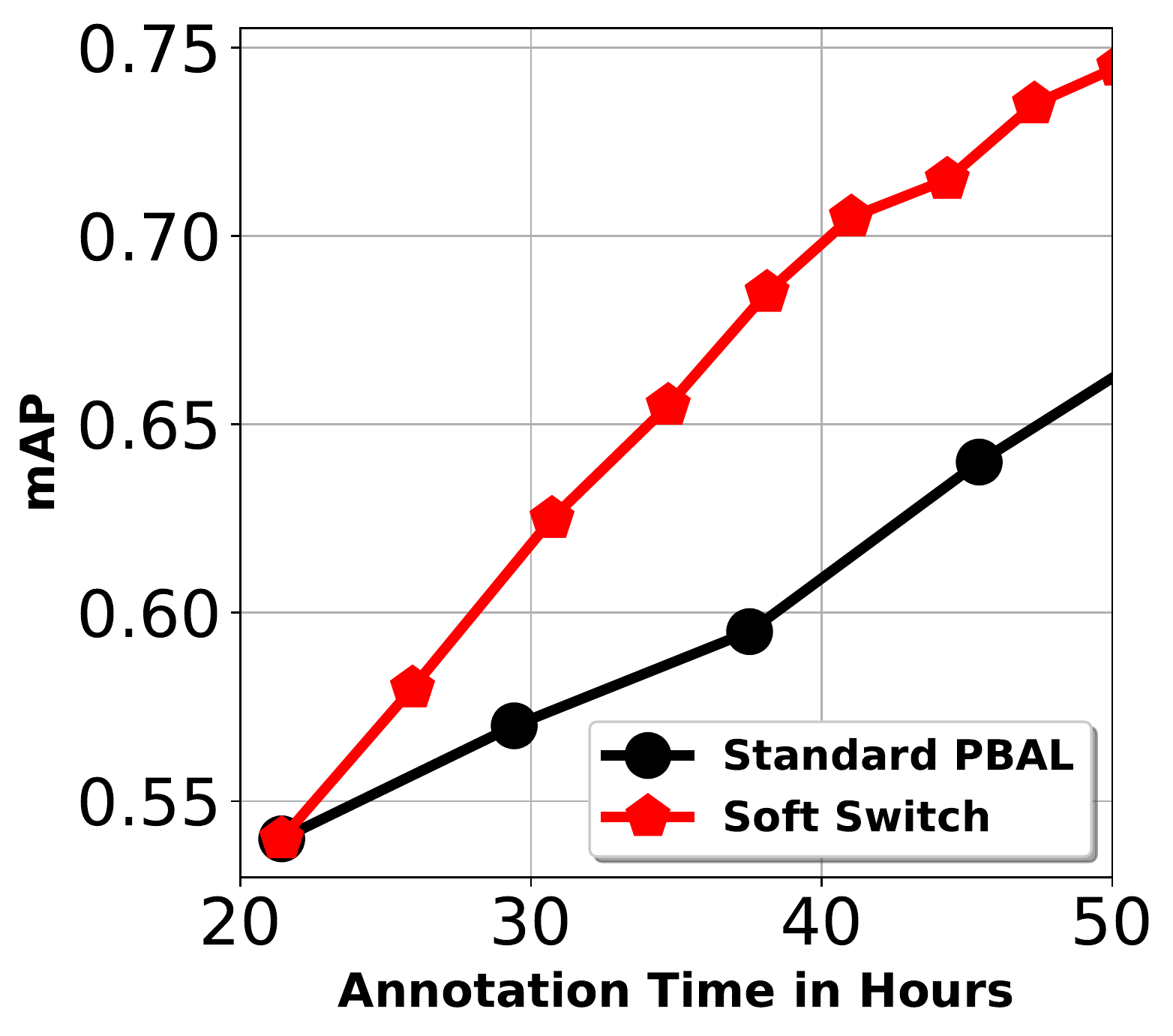}}\\
(a) Avg-Entropy &(b) Max-Margin & (c) Least-Confident
\end{tabular}
\caption{Wheat: For each active query method, we show performance of our adaptive supervision methods against standard PBAL framework (budget = 50 hours).}
\vspace{-15pt}
\label{fig:wheat_graphs}
\end{figure}

\noindent
\textbf{Evaluation.} Figure \ref{fig:wheat_graphs} shows the performance comparison of various supervision techniques for three different active sampling techniques. It can be observed that soft switch performs better than all other supervision techniques. As an example, for avg-entropy sampling (Figure \ref{fig:wheat_graphs}a), to attain a test mAP of around 0.68, soft switch method requires 34 hours of annotation (24\% savings), hard switch requires 38 hours (15.5\% savings) whereas the standard PBAL method requires around 45 hours.   

\vspace{-15pt}
\section{Discussion}
\vspace{-5pt}

Given an annotation budget, the choice of hard switch and soft switch thresholds is crucial in getting optimum performance out of the model. A higher value of $\gamma$ (hard switch threshold) results in a quicker switch to strong supervision, which would quickly deplete the annotation budget. Similarly, a lower $\gamma$ would result in a delayed switch to strong labeling. This would reduce strong label requirement but at the cost of providing a lot of noisy labels to the model. Similarly, a lower $\delta$ (soft switch threshold) would provide a lot of noisy labels to the model and a higher $\delta$ might mostly query only for strong labels. In other words, the hard switch and soft switch thresholds can be seen as knobs to adjust label quality, annotation costs and the number of training episodes taken to reach a desired level of performance.

\begin{wrapfigure}{r}{0.35\textwidth}
 \vspace{-7pt}
  \begin{center}
    \vspace{-4mm}
    \includegraphics[width=\linewidth]{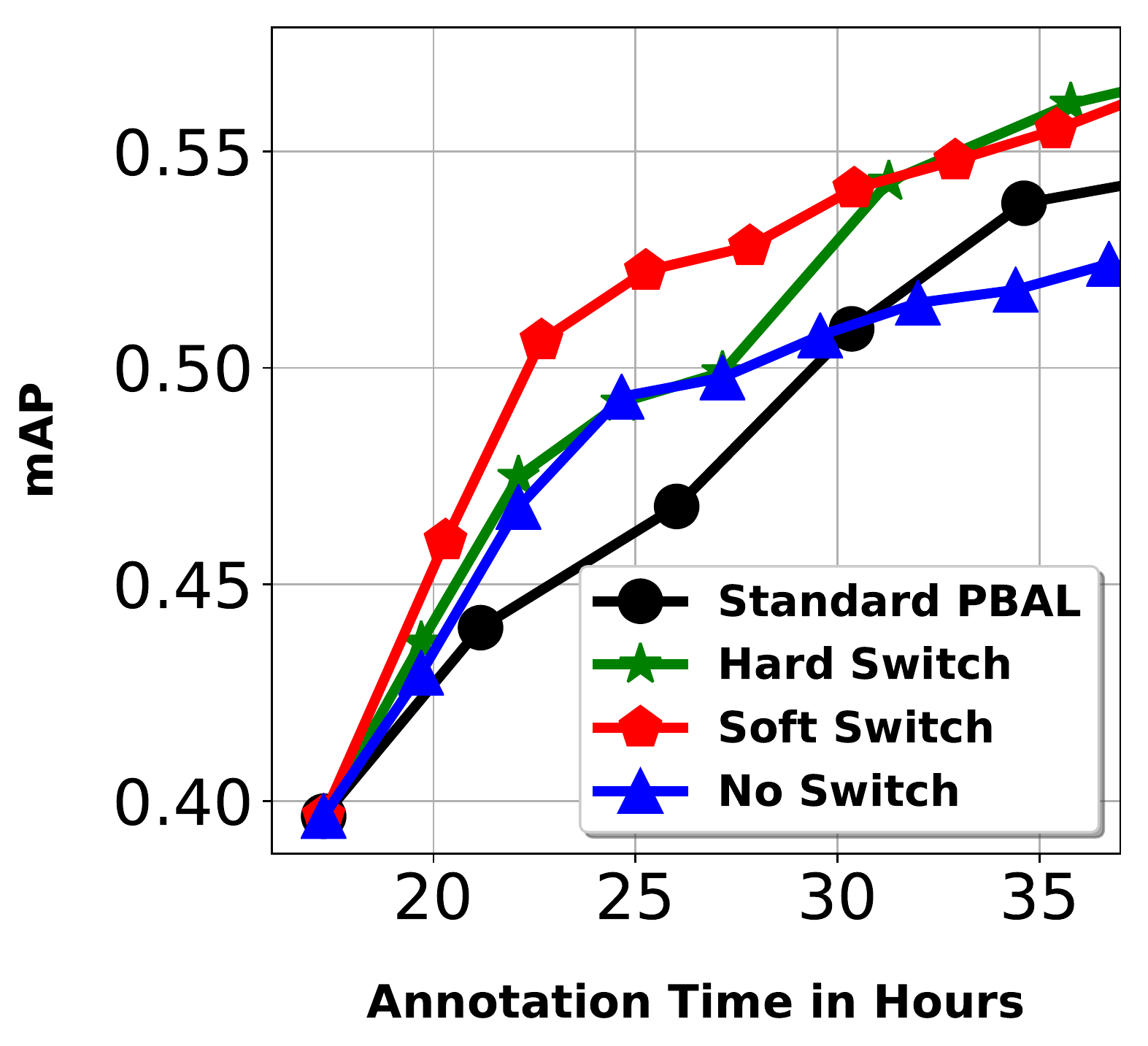}
  \end{center}
  \vspace{-6mm}
  \caption{\small Performance comparison of different supervision techniques in passive learning on PASCAL VOC 07}
  \vspace{-5pt}
  \label{fig:voc7_random}
\end{wrapfigure}

To understand the effect of adaptive supervision alone without active sampling, we conducted an ablation study to evaluate the performance of our framework in the context of a passive learning (random sampling) setting. It can be seen in Figure \ref{fig:voc7_random} that our adaptive supervision methods still outperform the standard PBAL method. To attain a test mAP of 0.53, standard PBAL requires around 35 hours of annotation time whereas hard switch requires 31 hours (11.4\% savings) and soft switch requires 30.4 hours (13.1\% savings). From this experiment, we observe that our methods reduce annotation effort even in the passive sampling case, albeit a lower percentage of savings when compared to performance on top of active sampling techniques. 



\vspace{-20pt}
\section{Conclusions}
\vspace{-5pt}
Using our proposed adaptive supervision framework, we empirically show that active learning approaches can be interleaved with multiple levels of supervision to achieve significant savings in annotation effort required to train deep object detectors. By only using the prediction outputs of the object detection model, we develop two supervision switching techniques: hard switch (inter episode switch) and soft switch (intra episode switch). Our experiments show that our adaptive supervision methods outperform standard PBAL on standard active query techniques. We believe that our work could open up a range of possibilities in fusing weak supervision techniques with active learning such as: using other forms of weak supervision with active learning, posing the problem of combining weak and strong supervision as an optimization problem under given budget constraints, combining active learning techniques with data programming based weak supervision techniques to name a few.  

\bibliography{egbib}
\end{document}